\documentclass[runningheads,a4paper]{llncs}
\usepackage{bbding}
\usepackage{amssymb}
\usepackage{graphicx}
\usepackage{amsmath}
\usepackage{longtable}
\usepackage{float}

\usepackage{url}
\urldef{\mailsa}\path|mohanlin@ict.ac.cn|
\newcommand{\keywords}[1]{\par\addvspace\baselineskip
\noindent\keywordname\enspace\ignorespaces#1}

\begin{document}
\mainmatter
\title{A Kind of Affine Weighted Moment Invariants}
\titlerunning{A Kind of Affine Weighted Moment Invariants}
\author{Hanlin Mo$^{1}$$^{(}$\Envelope$^{)}$$^{,}$%
\thanks{Student is the first author.}%
\and Shirui Li$^{1}$\and You Hao$^{1}$\and Hua Li$^{1}$}
\authorrunning{H.L. Mo et al.}
\institute{$^1$Key Laboratory of Intelligent Information Processing,\\Institute of Computing Technology, Chinese Academy of Sciences, Beijing, China\\
\mailsa\\}
\toctitle{Lecture Notes in Computer Science}
\tocauthor{Authors' Instructions}
\maketitle
\begin{abstract}
A new kind of geometric invariants is proposed in this paper, which is called affine weighted moment invariant (AWMI). By combination of local affine differential invariants and a  framework of global integral, they can more effectively extract features of images and help to increase the number of low-order invariants and to decrease the calculating cost. The experimental results show that AWMIs have good stability and distinguishability and achieve better results in image retrieval than traditional moment invariants. An extension to 3D is straightforward.
\keywords{differential invariants, integral invariants, weighted moment, affine transform, affine weighted moment invariants, low-order}
\end{abstract}
\section{Introduction}
Researchers have found that the geometric deformation of the object, caused by the change of viewpoint, is an important factor leading to the object to be misidentified. In order to solve this problem, various methods have been proposed in order to get image features which are robust to the geometric deformation. Moments and moment invariants are one of them.

The concepts of moment and moment invariants were first proposed by Hu in 1962 \cite{1}. He employed the theory of algebraic invariants, which was studied in 19th century\cite{2}, and defined geometric moments. Then he constructed seven geometric moment invariants which are invariant to the similarity transform. This set of invariants was widely used in various fields of pattern recognition, like \cite{3}. But the similarity transform can't represent all geometric deformations. When the distance between the camera and the object is much larger than the size of the object itself, the geometric deformation of the object can be represented by the affine transform. The landmark work of affine moment invariants(AMIs) was proposed by Flusser and Suk in 1993 \cite{4}. They used geometric moments to construct several low-order and low-degree AMIs, which were more effective to practical applications, for example, image registration \cite{5}. In order to obtain more AMIs, Suk and Flusser proposed the graph method which can generate AMIs of every order and degree \cite{6}. Xu and Li \cite{7} derived moment invariants in the intuitive way by multiple integrals of invariant geometric primitives like distance, area and volume. This method not only simplified the construction of AMIs, but also made them have a clear geometric meaning. Recently, Li $el\ at.$ improved the method of geometric primitives \cite{8}. They found a way to further simplify geometric primitives and used dot-product and cross-product of vectors to generate invariants. Meanwhile, researchers were also constantly expanding the definition of moments. Due to the lack of orthogonality, information redundancy of geometric moments became inevitable and image reconstruction from geometric moments was very difficult \cite{8}. Therefore, various orthogonal polynomials were used to define new moments. The first one proposed in 1980, Teague introduced orthogonal Legendre and Zernike moments \cite{9}. Then pseudo-Zernike moments \cite{10}, Fourier-Mellin moments \cite{11}, Chebyshey-Fourier moments \cite{12}, pseudo-Jacobi-Fourier moments \cite{13} and Gaussian Hermite moments \cite{14} were proposed. However, it was very difficult to obtain affine moment invariants of these orthogonal moments. This weakness greatly limited the use of orthogonal moments. Additionally, previous studies have shown that low-order and low-degree moment invariants have better performance, such as stability, than high-order and high-degree moment invariants. But the number of low-order and low-degree moment invariants was very limited. So, it’s very useful to get more low-order and low-degree moment invariants.

The studies of local differential invariants are another area, which need to be concerned. Olver generalized the moving frame method and got differential invariants for general transformation groups \cite{15}. He defined the affine gradient by using local affine differential invariants \cite{16}. Ge $et\ al.$ \cite{17} presented a local feature descriptor under color affine transformation by using the affine gradient. Wang $et\ al.$ \cite{18} proposed an effective method to derive a special type of affine differential invariants. Given some functions defined on the plane and affine group acting on the plane. However, they didn't explain how to use these local affine differential invariants in practical applications and how to improve the numerical accuracy of partial derivatives on discrete image.

In this paper, we use the frame of geometric moments and partial derivatives to define a kind of weighted moments, which can be named as differential moments(DMs). According to the definition of DMs and local affine differential invariants, affine weighted moment invariants(AWMIs) can be obtained easily, which use both global and local information. The experimental results show that AWMIs have good stability and distinguishability. Also, they can improve the accuracy of image retrieval.

\section{Some Basic Definitions and Theorems}
In order to understand the construction frame of AWMIs more clearly, we first introduce some basic definitions and theorems.
\subsection{The Definition of Geometric Moments}
The geometric moment of the image $f(x,y)$ is defined by
\begin{equation}
m_{pq}
=
\int_{\infty}^{-\infty}\int_{\infty}^{-\infty}x^{p}y^{q}f(x,y)dxdy
\end{equation}
where $p$,$q$ $=$ $\{0,1,2,...\}$, $(p+q)$ is the order of $m_{pq}$. In order to eliminate the effect of translation, central geometric moments are usually used. The central moment of the order $(p+q)$ is defined by
\begin{equation}
u_{pq}
=
\int_{\infty}^{-\infty}\int_{\infty}^{-\infty}(x-\bar x)^{p}(y-\bar y)^{q}f(x,y)dxdy
\end{equation}
where
\begin{equation}
 \bar x =\frac {m_{10}}{m_{00}},~~\bar y =\frac {m_{01}}{m_{00}}
\end{equation}
\subsection{Coordinate Transformation under the Affine Transform}
Suppose the image $f(x,y)$ is transformed into another image $g(x^{'},y^{'})$ by the affine transform A and the translation T. $(x^{'},y^{'})$ is the corresponding point of $(x,y)$. We can get the following relationship
\begin{equation}
\left(
\begin{array}{c}
  x^{'}\\
  y^{'}\\
\end{array}
\right)
=
A
\cdot
\left(
\begin{array}{c}
  x\\
  y\\
\end{array}
\right)
+
T
=
\left(
\begin{array}{cc}
  a_{11}&a_{12}\\
  a_{21}&a_{22}\\
\end{array}
\right)
\cdot
\left(
\begin{array}{c}
  x\\
  y\\
\end{array}
\right)
+
\left(
\begin{array}{c}
  t_1\\
  t_2\\
\end{array}
\right)
\end{equation}
where A is a nonsingular matrix.

\subsection{The Construction of Affine Moments Invariants}
For the image $f(x,y)$, let $(x_{i},y_{i})$ and $(x_{j},y_{j})$ be two arbitrary points in the domain of $f(x,y)$. The geometric primitive proposed in \cite{7} can be defined by
\begin{equation}
S(i,j)=
\left|
\begin{array}{cc}
  (x_{i}-\bar x)&(x_{j}-\bar x)\\
  (y_{i}-\bar y)&(y_{j}-\bar y)\\
\end{array}
\right|
\end{equation}

Suppose the image $f(x,y)$ is transformed into another image $g(x^{'},y^{'})$ by Eq.(4). $(x^{'}_{i},y^{'}_{i}), (x^{'}_{j},y^{'}_{j})$ in $g(x^{'},y^{'})$ are the corresponding points of $(x_{i},y_{i}), x_{j},y_{j})$ in $f(x,y)$. Then, there is a relation
\begin{equation}
S^{'}(i,j)=|A|\cdot S(i,j)
\end{equation}
where $|A|$ is the determinant of A. Therefore, using N points $(x_{1},y_{1}),(x_{2},y_{2}),...,(x_{N},y_{N})$ in $f(x,y)$ , $Core(N,m;d_{1},d_{2},....,d_{N})$ can be defined by
\begin{equation}
Core(N,m;d_{1},d_{2},...,d_{N})=
\underbrace{S(o,1,2)...S(o,k,l)...S(o,r,N)}_m
\end{equation}
where $k<l$, $r<N$,\ $k,l,r$ $\in$
$\{1,2,...,N \}$.
$d_{i}$ represents the number of point $(x_{i},y_{i})$ in all geometric primitives, $i=1,2,...,N$.

Let $(x^{'}_{1},y^{'}_{1}),(x^{'}_{2},y^{'}_{2}),...,(x^{'}_{N},y^{'}_{N})$ in $g(x^{'},y^{'})$ be corresponding points of $(x_{1},y_{1}),\\(x_{2},y_{2}),...,(x_{N},y_{N})$ in $f(x,y)$. It's obviously that
\begin{equation}
Core^{'}(N,m;d_{1},d_{2},....,d_{N})=|A|^{m}Core(N,m;d_{1},d_{2},....,d_{N})
\end{equation}
where
\begin{equation}
Core^{'}(N,m;d_{1},d_{2},...,d_{N})=
\underbrace{S^{'}(o,1,2)...S^{'}(o,k,l)...S^{'}(o,r,N)}_m
\end{equation}
Finally, using $Core(N,m;d_{1},d_{2},....,d_{N})$, $AMIs$ can be defined by
\begin{equation}
\begin{split}
AMIs=&\frac {I(Core(N,m;d_{1},d_{2},....,d_{N}))}{(\iint f(x,y) dxdy)^{N+m}}\\
&=\frac {\iint ... \iint Core(N,m;d_{1},d_{2},....,d_{N})\, dx_{1}dy_{1}\,...\,dx_{N}dy_{N}}{(\iint f(x,y) dxdy)^{N+m}}
\end{split}
\end{equation}

In \cite{7}, Xu and Li proved that Eq.(10) didn't change when the image was transformed by Eq.(4). Eq.(10) is the general form of AMIs. In fact, this multiple integral can be expressed as polynomials of central geometric moments.
\begin{equation}
\frac {I(Core(N,m;d_{1},d_{2},....,d_{N}))}{(\iint f(x,y) dxdy)^{N+m}}=
\frac {\sum_{j}\limits a_{j} \cdot \prod_{i=1}^N \limits u_{p_{i}q_{i}}}{(u_{00})^{N+m}}
\end{equation}
where $j$ represents the number of multiplicative items in this expansion, $a_{j}$ represents the coefficient of the j-th multiplicative item. In general, $N$ is named as the degree of Eq.(10),\ $ \max \limits_{i} \left\{ p_{i}+q_{i} \right\} $ is named as the order of Eq.(10). They are determined by $Core(N,m;d_{1},d_{2},....,d_{N})$.

\subsection{Local Differential Invariants under the Affine Transform}
For the differentiable function $f(x,y)$, Olver \cite{15} used the contact-invariant coframe to obtain local differential invariants under the affine transform. The first and second order local differential invariants of $f(x,y)$ were defined by:
\begin{equation}
ADI_{1}=x\frac {\partial f}{\partial x}+y\frac {\partial f}{\partial y}
\end{equation}
\begin{equation}
ADI_{2}=x^{2}\frac {\partial f^{2}}{\partial^{2} x}+2xy\frac {\partial f^{2}}{\partial x \partial y}+y^{2}\frac {\partial f^{2}}{\partial^{2} y}
\end{equation}
\begin{equation}
ADI_{3}=x\frac {\partial f}{\partial y}\frac {\partial f^{2}}{\partial^{2} x}+(y\frac {\partial f}{\partial y}-x\frac {\partial f}{\partial x})\frac {\partial f^{2}}{\partial x \partial y}-y\frac {\partial f}{\partial x}\frac {\partial f^{2}}{\partial^{2} y}
\end{equation}
\begin{equation}
ADI_{4}=\frac {\partial f^{2}}{\partial^{2} x}\frac {\partial f^{2}}{\partial^{2} y}-(\frac {\partial f^{2}}{\partial x \partial y})^{2}
\end{equation}
\begin{equation}
ADI_{5}=(\frac {\partial f}{\partial y})^{2}\frac {\partial f^{2}}{\partial^{2} x}-2\frac {\partial f}{\partial x}\frac {\partial f}{\partial y}\frac {\partial f^{2}}{\partial x \partial y}+(\frac {\partial f}{\partial x})^{2}\frac {\partial f^{2}}{\partial^{2} y}
\end{equation}

Among them, $ADI_{4}$ and $ADI_{5}$ are pure differential invariants, which don't contain $x$ or $y$. $ADI_{1}$ and $ADI_{2}$ are absolute differential invariants. \ $ADI_{3}$ , $ADI_{4}$ and $ADI_{5}$ are relative differential invariants, which meanS
\begin{equation}
ADI_{3}=\frac {1}{|A|} ADI_{3}^{'}~~~~~~~ADI_{4}=\frac {1}{|A|^{2}} ADI_{4}^{'}~~~~~~~ADI_{5}=\frac {1}{|A|^{2}} ADI_{5}^{'}
\end{equation}
where $ADI_{1}^{'}, ADI_{2}^{'}, ADI_{3}^{'}, ADI_{4}^{'}$ and $ADI_{5}^{'}$ are local differential invariants of $g(x^{'},y^{'})$. $f(x,y)$ and $g(x^{'},y^{'})$ satisfy the relationship shown in Eq.(4).

In addition, Olver indicated that differential invariants shown in $(12) \sim(16)$ were not independent \cite{15}. The relationship of them was defined by
\begin{equation}
ADI_{2}^{2}-ADI_{5}ADI_{3}+ADI_{1}^{2}ADI_{4}=0
\end{equation}
\section{The Construction Frame of AWMIs}
\subsection{The Definition of DMs}
\textbf{Definition 1.}

Let $f(x,y)$ be the differentiable function. The first-order DMs are defined by:
\begin{equation}
D^{pq}_{mn}=
\int_{\infty}^{-\infty}\int_{\infty}^{-\infty}(x-\bar x)^{p}(y-\bar y)^{q}(\frac {\partial f}{\partial x})^{m}(\frac {\partial f}{\partial x})^{n}f(x,y) dxdy
\end{equation}
where $p,q,m,n$ $\in$ $N$.

The second-order differential moments are defined by:
\begin{equation}
\begin{split}
D^{pq}_{mnrst}&=
\int_{\infty}^{-\infty}\int_{\infty}^{-\infty}(x-\bar x)^{p}(y-\bar y)^{q}(\frac {\partial f}{\partial x})^{m}(\frac {\partial f}{\partial x})^{n}\\
&(\frac {\partial f^{2}}{\partial^{2} x})^r(\frac {\partial f^{2}}{\partial^{2} y})^s
(\frac {\partial f^{2}}{\partial x \partial y})^tf(x,y) dxdy \\
\end{split}
\end{equation}
where $p,q,m,n,r,s,t$ $\in$ $N$.

Similarly, we can construct higher-order differential moments. But considering their convenience and the accuracy of calculation, we only define the first-order and second-order DMs. Compared with the definition of geometric central moments in Eq(2), DMs are constructed by using the polynomial functions and derivative functions of $f(x,y)$. Thus, they can represent internal information of images better.
\subsection{The Construction of AWMIs}
\textbf{Definition 2.}

Suppose $f(x,y)$ be the differentiable function, the first-order AWMIs constructed by the the first-order DMs are defined by
\begin{equation}
\begin{split}
\hspace{-3mm}
ADMI_{1}=&\frac {I(DCore(N,m;d_{1},d_{2},....,d_{N},k_{1},k_{2},....,k_{N}))}{(\iint f(x,y) dxdy)^{N+m}} \\
=&\frac {\sum_{j}\limits a_{j} \cdot \prod_{i=1}^N \limits D^{p_{i}q_{i}}_{m_{i}n_{i}}}{(D^{00}_{00})^{N+m}}
\end{split}
\end{equation}
where
\begin{equation}
\begin{split}
&DCore(N,m;d_{1},d_{2},....,d_{N},k_{1},k_{2},....,k_{N})\\
&=Core(N,m;d_{1},d_{2},....,d_{N})(ADI_{1}^{1})^{k_{1}}(ADI_{1}^{2})^{k_{2}}...(ADI_{1}^{N}))^{k_{N}}\\
&=Core(N,m;d_{1},d_{2},....,d_{N})(x_{1}\frac {\partial f}{\partial x_{1}}+y_{1}\frac {\partial f}{\partial y_{1}})^{k_{1}}
(x_{2}\frac {\partial f}{\partial x_{2}}+y_{2}\frac {\partial f}{\partial y_{2}})^{k_{2}}\\
&...(x_{N}\frac {\partial f}{\partial x_{N}}+y_{N}\frac {\partial f}{\partial y_{N}})^{k_{N}}
\end{split}
\end{equation}

Note that we assume $x_{i}=x_{i}-\bar{x}$ and $y_{i}=y_{i}-\bar{y}\ (i=1,2,...,N)$. Then, we can get the following theorem.
\\

\hspace{-8mm}
\textbf{Theorem 1.}

Suppose the image $f(x,y)$ is transformed into another image $g(x^{'},y^{'})$ by Eq.(4). $(x^{'}_{1},y^{'}_{1}),(x^{'}_{2},y^{'}_{2}),...,(x^{'}_{N},y^{'}_{N})$ in $g(x^{'},y^{'})$ are corresponding points of $(x_{1},y_{1}),\\(x_{2},y_{2}),...,(x_{N},y_{N})$ in $f(x,y)$. The following equation are established.
\begin{equation}
\begin{split}
&\frac {I(DCore(N,m;d_{1},d_{2},....,d_{N},k_{1},k_{2},....,k_{N}))}{(\iint f(x,y) dxdy)^{N+m}}
\\=&\frac {I(DCore^{'}(N,m;d_{1},d_{2},....,d_{N},k_{1},k_{2},....,k_{N}))}{(\iint g(x^{'},y^{'}) dx^{'}dy^{'})^{N+m}}\\
\end{split}
\end{equation}
where
\begin{equation}
\begin{split}
&DCore^{'}(N,m;d_{1},d_{2},....,d_{N},k_{1},k_{2},....,k_{N})\\
&=Core^{'}(N,m;d_{1},d_{2},....,d_{N})(ADI_{1}^{1})^{k_{1}}(ADI_{1}^{2})^{k_{2}}...(ADI_{1}^{N}))^{k_{N}}\\
&=Core^{'}(N,m;d_{1},d_{2},....,d_{N})(x^{'}_{1}\frac {\partial g}{\partial x^{'}_{1}}+y^{'}_{1}\frac {\partial g}{\partial y^{'}_{1}})^{k_{1}}
(x^{'}_{2}\frac {\partial g}{\partial x^{'}_{2}}+y^{'}_{2}\frac {\partial g}{\partial y^{'}_{2}})^{k_{2}}\\
&...(x^{'}_{N}\frac {\partial g}{\partial x^{'}_{N}}+y^{'}_{N}\frac {\partial g}{\partial y^{'}_{N}})^{k_{N}}
\end{split}
\end{equation}

The proof of Eq.(23) is the same as that of Eq.(10) proved in \cite{7} .
\subsection{The Instances of AWMIs}
In \cite{6}, Flusser and Suk proved that there were seven kinds of AMIs when the degree $N$  $\leqslant 3$ and the order $ \max \limits_{i} \left\{ p_{i}+q_{i} \right\} $  $\leqslant 3$. They are listed in the Table 1. It is important to note that $u_{10}$ and $u_{01}$ are always zero. Thus, $AMI4$ and $AMI5$ can't be used as invariants. That means there are only two AMIs, \{$AMI2, AMI7$\}, when the degree $N \leqslant 3$ and the order $ \max \limits_{i} \left\{ p_{i}+q_{i} \right\} $  $\leqslant 3$.

\begin{table}[H] 
\centering
\caption{AMIs ($N \leqslant 3$, $ \max \limits_{i} \left\{ p_{i}+q_{i} \right\} $  $\leqslant 3$)} 
\begin{tabular}{p{1cm}|p{5.5cm}|p{6cm}} 
\hline
No. & Core & AMI \\ 
\hline 
$AMI1$ & $(x_{1}y_{2}-x_{2}y_{1})$& 0\\
$AMI2$ & $(x_{1}y_{2}-x_{2}y_{1})^{2}$& $2u_{02}u_{20}-2u_{11}^{2}$\\
$AMI3$ & $(x_{1}y_{2}-x_{2}y_{1})^{3}$& 0\\
$AMI4$ & $(x_{1}y_{2}-x_{2}y_{1})(x_{1}y_{3}-x_{3}y_{1})$& $u_{01}^{2}u_{20}-2u_{01}u_{10}u_{11}+u_{02}u_{10}^2$\\
$AMI5$ & $(x_{1}y_{2}-x_{2}y_{1})(x_{1}y_{3}-x_{3}y_{1})^{2}$& $u_{01}u_{02}u_{30}-2u_{01}u_{11}u_{21}+u_{01}u_{12}u_{20}-u_{02}u_{10}u_{21}-u_{03}u_{10}u_{20}+2u_{10}u_{11}u_{12}$\\
$AMI6$ & $(x_{1}y_{2}-x_{2}y_{1})(x_{1}y_{3}-x_{3}y_{1})(x_{2}y_{3}-x_{3}y_{2})$& 0\\
$AMI7$ & $(x_{1}y_{2}-x_{2}y_{1})(x_{1}y_{3}-x_{3}y_{1})(x_{2}y_{3}-x_{3}y_{2})^{2}$& $2u_{02}u_{12}u_{30}-2u_{02}u_{21}^{2}-2u_{03}u_{11}u_{30}+2u_{03}u_{20}u_{21}+2u_{11}u_{12}u_{21}-2u_{12}^{2}u_{20}$\\
\hline
\end{tabular}
\end{table}

But now, we can use Eq.(21) to construct many AWMIs. When $N \leqslant 3$, $ \max \limits_{i} \left\{ p_{i}+q_{i} \right\} $  $\leqslant 3$ and $ \max \limits_{i} \left\{ m_{i}+n_{i} \right\} $  $\leqslant 1$, there are 8 kinds of DCores which can be constructed AWMIs. They are list in the Table 2.

\begin{table}[H]
\caption{{DCores} ($N \leqslant 3$, $ \max \limits_{i} \left\{ p_{i}+q_{i} \right\} $  $\leqslant 3$, $\max \limits_{i} \left\{ m_{i}+n_{i} \right\} $  $\leqslant 1$)} 
\begin{tabular}{p{1.5cm}|p{11cm}} 
\hline
No. & DCore \\ 
\hline 
$DCore_{1}$  & $(x_{1}y_{2}-x_{2}y_{1})^{2}(x_{1}\frac {\partial f}{\partial x_{1}}+y_{1}\frac {\partial f}{\partial y_{1}})$ \\
$DCore_{2}$  & $(x_{1}y_{2}-x_{2}y_{1})^{2}(x_{1}\frac {\partial f}{\partial x_{1}}+y_{1}\frac {\partial f}{\partial y_{1}})(x_{2}\frac {\partial f}{\partial x_{2}}+y_{2}\frac {\partial f}{\partial y_{2}})$ \\
$DCore_{3}$ & $(x_{1}y_{2}-x_{2}y_{1})(x_{1}y_{3}-x_{3}y_{1})(x_{2}\frac {\partial f}{\partial x_{2}}+y_{2}\frac {\partial f}{\partial y_{2}})(x_{3}\frac {\partial f}{\partial x_{3}}+y_{3}\frac {\partial f}{\partial y_{3}})$ \\
$DCore_{4}$ & $(x_{1}y_{2}-x_{2}y_{1})(x_{1}y_{3}-x_{3}y_{1})(x_{1}\frac {\partial f}{\partial x_{1}}+y_{1}\frac {\partial f}{\partial y_{1}})(x_{2}\frac {\partial f}{\partial x_{2}}+y_{2}\frac {\partial f}{\partial y_{2}})(x_{3}\frac {\partial f}{\partial x_{3}}+y_{3}\frac {\partial f}{\partial y_{3}})$ \\
$DCore_{5}$ & $(x_{1}y_{2}-x_{2}y_{1})(x_{1}y_{3}-x_{3}y_{1})^{2}(x_{2}\frac {\partial f}{\partial x_{2}}+y_{2}\frac {\partial f}{\partial y_{2}})$ \\
$DCore_{6}$ & $(x_{1}y_{2}-x_{2}y_{1})(x_{1}y_{3}-x_{3}y_{1})^{2}(x_{3}\frac {\partial f}{\partial x_{3}}+y_{3}\frac {\partial f}{\partial y_{3}})$ \\
$DCore_{7}$ & $(x_{1}y_{2}-x_{2}y_{1})(x_{1}y_{3}-x_{3}y_{1})^{2}(x_{2}\frac {\partial f}{\partial x_{2}}+y_{2}\frac {\partial f}{\partial y_{2}})(x_{3}\frac {\partial f}{\partial x_{3}}+y_{3}\frac {\partial f}{\partial y_{3}})$ \\
$DCore_{8}$ & $(x_{1}y_{2}-x_{2}y_{1})(x_{1}y_{3}-x_{3}y_{1})(x_{2}y_{3}-x_{3}y_{2})^{2}(x_{1}\frac {\partial f}{\partial x_{1}}+y_{1}\frac {\partial f}{\partial y_{1}})$ \\
\hline
\end{tabular}
\end{table}
In the Table 3, we list AWMIs constructed by Dcores in the Table 2. They are all constructed by first-order DMs. It is worth noting that we have removed Dcores of which expansions are always 0 or contain $D^{10}_{00}, D^{01}_{00}$. In fact, using a similar definition to Eq.(23), we can get AWMIs constructed by the second-order DMs. But here, we give a new definition. We want to point out that there are many different methods to construct AWMIs.
\begin{table}[H]
\caption{{AWMIs} ($N \leqslant 3$, $ \max \limits_{i} \left\{ p_{i}+q_{i} \right\} $  $\leqslant 3$, $\max \limits_{i} \left\{ m_{i}+n_{i} \right\} $  $\leqslant 1$)}
\begin{tabular}{p{1.5cm}|p{10cm}}
\hline
No. & AWMI \\ 
\hline 
&\\
$AWMI^{1}_{1}$  & $D^{02}_{00}D^{21}_{01}+D^{02}_{00}D^{30}_{10}+D^{03}_{01}D^{20}_{00}-2D^{11}_{00}D^{12}_{01}-2D^{11}_{00}D^{21}_{10}+D^{12}_{10}D^{20}_{00}$\\
&\\
$AWMI^{2}_{1}$  &  $D^{03}_{01}D^{21}_{01}+D^{03}_{01}D^{30}_{10}-(D^{12}_{01})^{2}-2D^{12}_{01}D^{21}_{10}+D^{12}_{10}D^{21}_{01}+D^{12}_{10}D^{30}_{10}-(D^{21}_{10})^{2}$\\
&\\
$AWMI^{3}_{1}$ &
$D^{02}_{00}(D^{11}_{01})^{2}+2D^{02}_{00}D^{11}_{01}D^{20}_{10}+D^{02}_{00}(D^{20}_{10})^{2}+(D^{02}_{01})^{2}D^{20}_{00}-2D^{02}_{01}D^{11}_{00}D^{11}_{01}-2D^{02}_{01}D^{11}_{00}D^{20}_{10}+2D^{02}_{01}D^{11}_{10}D^{20}_{00}-2D^{11}_{00}D^{11}_{01}D^{11}_{10}-2D^{11}_{00}D^{11}_{10}D^{20}_{10}+(D^{11}_{10})^{2}D^{20}_{00}$\\
&\\
$AWMI^{4}_{1}$ &
$(D^{02}_{01})^{2}D^{21}_{01}+(D^{02}_{01})^{2}D^{30}_{10}-2D^{02}_{01}D^{11}_{01}D^{12}_{01}-2D^{02}_{01}D^{11}_{01}D^{21}_{10}+2D^{02}_{01}D^{11}_{10}D^{21}_{01}+2D^{02}_{01}D^{11}_{10}D^{30}_{10}-2D^{02}_{01}D^{12}_{01}D^{20}_{10}-2D^{02}_{01}D^{20}_{10}D^{21}_{10}+D^{03}_{01}(D^{11}_{01})^{2}+2D^{03}_{01}D^{11}_{01}D^{20}_{10}+D^{03}_{01}(D^{20}_{10})^{2}+(D^{11}_{01})^{2}D^{12}_{10}-2D^{11}_{01}D^{11}_{10}D^{12}_{01}-2D^{11}_{01}D^{11}_{10}D^{21}_{10}+2D^{11}_{01}D^{12}_{10}D^{20}_{10}+(D^{11}_{10})^{2}D^{21}_{01}+(D^{11}_{10})^{2}D^{30}_{10}-2D^{11}_{10}D^{12}_{01}D^{20}_{10}-2D^{11}_{10}D^{20}_{10}D^{21}_{10}+D^{12}_{10}(D^{20}_{10})^{2}$\\
&\\
$AWMI^{5}_{1}$ &
$D^{02}_{00}D^{02}_{01}D^{30}_{00}-D^{02}_{00}D^{11}_{01}D^{21}_{00}+D^{02}_{00}D^{11}_{10}D^{30}_{00}-D^{02}_{00}D^{20}_{10}D^{21}_{00}-2D^{02}_{01}D^{11}_{00}D^{21}_{00}+D^{02}_{01}D^{12}_{00}D^{20}_{00}-D^{03}_{00}D^{11}_{01}D^{20}_{00}-D^{03}_{00}D^{20}_{00}D^{20}_{10}+2D^{11}_{00}D^{11}_{01}D^{12}_{00}-2D^{11}_{00}D^{11}_{10}D^{21}_{00}+2D^{11}_{00}D^{12}_{00}D^{20}_{10}+D^{11}_{10}D^{12}_{00}D^{20}_{00}$\\
&\\
$AWMI^{6}_{1}$ &
$D^{02}_{00}(D^{11}_{01})^{2}+2D^{02}_{00}D^{11}_{01}D^{20}_{10}+D^{02}_{00}(D^{20}_{10})^{2}+(D^{02}_{01})^{2}D^{20}_{00}-2D^{02}_{01}D^{11}_{00}D^{11}_{01}-2D^{02}_{01}D^{11}_{00}D^{20}_{10}+2D^{02}_{01}D^{11}_{10}D^{20}_{00}-2D^{11}_{00}D^{11}_{01}D^{11}_{10}-2D^{11}_{00}D^{11}_{10}D^{20}_{10}+(D^{11}_{10})^{2}D^{20}_{00}$\\
&\\
$AWMI^{7}_{1}$ &
$D^{02}_{01}D^{03}_{01}D^{30}_{00}+D^{02}_{01}D^{12}_{00}D^{21}_{01}+D^{02}_{01}D^{12}_{00}D^{30}_{10}-2D^{02}_{01}D^{12}_{01}D^{21}_{00}+D^{02}_{01}D^{12}_{10}D^{30}_{00}-2D^{02}_{01}D^{21}_{00}D^{21}_{10}-D^{03}_{00}D^{11}_{01}D^{21}_{01}-D^{03}_{00}D^{11}_{01}D^{30}_{10}-D^{03}_{00}D^{20}_{10}D^{21}_{01}-D^{03}_{00}D^{20}_{10}D^{30}_{10}-D^{03}_{01}D^{11}_{01}D^{21}_{00}+D^{03}_{01}D^{11}_{10}D^{30}_{00}-D^{03}_{01}D^{20}_{10}D^{21}_{00}+2D^{11}_{01}D^{12}_{00}D^{12}_{01}+2D^{11}_{01}D^{12}_{00}D^{21}_{10}-D^{11}_{01}D^{12}_{10}D^{21}_{00}+D^{11}_{10}D^{12}_{00}D^{21}_{01}+D^{11}_{10}D^{12}_{00}D^{30}_{10}-2D^{11}_{10}D^{12}_{01}D^{21}_{00}+D^{11}_{10}D^{12}_{10}D^{30}_{00}-2D^{11}_{10}D^{21}_{00}D^{21}_{10}+2D^{12}_{00}D^{12}_{01}D^{20}_{10}+2D^{12}_{00}D^{20}_{10}D^{21}_{10}-D^{12}_{10}D^{20}_{10}D^{21}_{00}$\\
&\\
$AWMI^{8}_{1}$ &
$-2D^{03}_{00}D^{12}_{01}D^{30}_{00}+2D^{03}_{00}D^{21}_{00}D^{21}_{01}+2D^{03}_{00}D^{21}_{00}D^{30}_{10}-2D^{03}_{00}D^{21}_{10}D^{30}_{00}+2D^{03}_{01}D^{12}_{00}D^{30}_{00}-2D^{03}_{01}(D^{21}_{00})^{2}-2(D^{12}_{00})^{2}D^{21}_{01}-2(D^{12}_{00})^{2}D^{30}_{10}+2D^{12}_{00}D^{12}_{01}D^{21}_{00}+2D^{12}_{00}D^{12}_{10}D^{30}_{00}+2D^{12}_{00}D^{21}_{00}D^{21}_{10}-2D^{12}_{10}(D^{21}_{00})^{2}$\\
&\\
\hline
\end{tabular}
\end{table}

\hspace{-8mm}
\textbf{Definition 3.}

Suppose $f(x,y)$ is the differentiable function, its AWMIs which are constructed by the second-order DMs can be defined by
\begin{equation}
AWMI_{2}=\frac {\iint ADI_{4}f(x,y) dxdy}{\iint ADI_{5}f(x,y) dxdy}
\end{equation}

According to Eq.(15) and Eq.(16), we can prove that Eq.(30) won't change when $f(x,y)$ is transformed by Eq.(4) very easily. Its expansion is defined by
\begin{equation}
AWMI_{2}=\frac {D^{00}_{00110}-D^{00}_{00002}}{D^{00}_{02100}-2D^{00}_{11001}+D^{00}_{20010}}
\end{equation}
\subsection{Differentials of Digital Images}
In the above, we assume that the function $f(x,y)$ is continuous and differentiable. Actually, general images are discrete two-dimensional functions. So we have to choose a way to calculate differentials more accurately. Some researchers have confirmed that employing derivatives of the Gaussian function as filters to compute derivatives of discrete functions via convolution is a good way \cite{19}.
The two-dimensional zeros-mean Gaussian functions and its the first-order and second-order differentials are defined by
\begin{equation}
\begin{split}
&G(x,y)=\frac {1}{2\pi\sigma^{2}}\mathrm{e}^{-\frac{x^{2}+y^{2}}{2\sigma^{2}}}~~~~~\frac {\partial G}{\partial x}=-\frac{x}{2\pi\sigma^{4}}\mathrm{e}^{-\frac{x^{2}+y^{2}}{2\sigma^{2}}}\\
&\frac {\partial G}{\partial y}=-\frac{y}{2\pi\sigma^{4}}\mathrm{e}^{-\frac{x^{2}+y^{2}}{2\sigma^{2}}}
~~~~~\frac {\partial^{2} G}{\partial x^{2}}=\frac{(x^{2}-\sigma^{2})}{2\pi\sigma^{6}}\mathrm{e}^{-\frac{x^{2}+y^{2}}{2\sigma^{2}}}\\
&\frac {\partial^{2} G}{\partial x \partial y}=\frac{xy}{2\pi\sigma^{6}}\mathrm{e}^{-\frac{x^{2}+y^{2}}{2\sigma^{2}}}
~~~~~\frac {\partial^{2} G}{\partial y^{2}}=\frac{(y^{2}-\sigma^{2})}{2\pi\sigma^{6}}\mathrm{e}^{-\frac{x^{2}+y^{2}}{2\sigma^{2}}}
\end{split}
\end{equation}

where $\sigma$ is the standard deviation.

Using Eq.(33)$\sim$(37) to be convolved with the image function $f(x,y)$, we can get partial derivatives of $f(x,y)$.
For example,
\begin{equation}
\frac {\partial f}{\partial x}=\frac {\partial G}{\partial x}*f(x,y)
\end{equation}
where $*$ means convolution. In this paper, we make $\sigma=3.0$ and kernel size $9\times9$.

\section{Experimental results and analysis}
In this section, some experiments are provided to evaluate the theoretical framework proposed in the previous sections. We will test the performance of AWMIs. In the first subsection, we calculate AWMIs of synthetic images to verify the stability and discernibility. The second subsection, we test the retrieval ability of AWMIs on the real image dataset. At the same time, we compare our AWMIS with several traditional moment invariants.
\subsection{Numerical stability and discernibility of AWMIs}
We choose 5 kinds of fish pictures from Web page:
\url{https://www.igfa.org/Fish/Fish-Database.aspx}. Original images are transformed by 5 different affine transformations and translations in Table 4.
\begin{table}[H]
\centering
\caption{\ Affine transformations}
\begin{tabular}{p{1cm}|p{1cm}|p{1cm}|p{1cm}|p{1cm}|p{1cm}|p{1cm}} 
\hline
No. & $a_{11}$ & $a_{12}$ & $a_{21}$ & $a_{22}$ & $t_{1}$ & $t_{2}$ \\ 
\hline 
1 & 0.69& -0.12 & 0.21 &1.18 &0 &150\\
2 & 0.57& 0.42 & -0.42 &0.42 &160 &280\\
3 & 0.60& -1.03 & 0.52 &0.30 &50 &15\\
4 & 1.00& -1.00 & 0.00 &1.00 &100 &50\\
5 & 1.50& 0.00 & 0.00 &0.80 &30 &10\\
\hline
\end{tabular}
\end{table}
Thus, 36 images are obtained($512 \times 512$), which are shown in the Figure.1. They can be divided into 6 groups, each group contains 6 images.
\begin{figure}
  \centering
  \includegraphics[width=10cm]{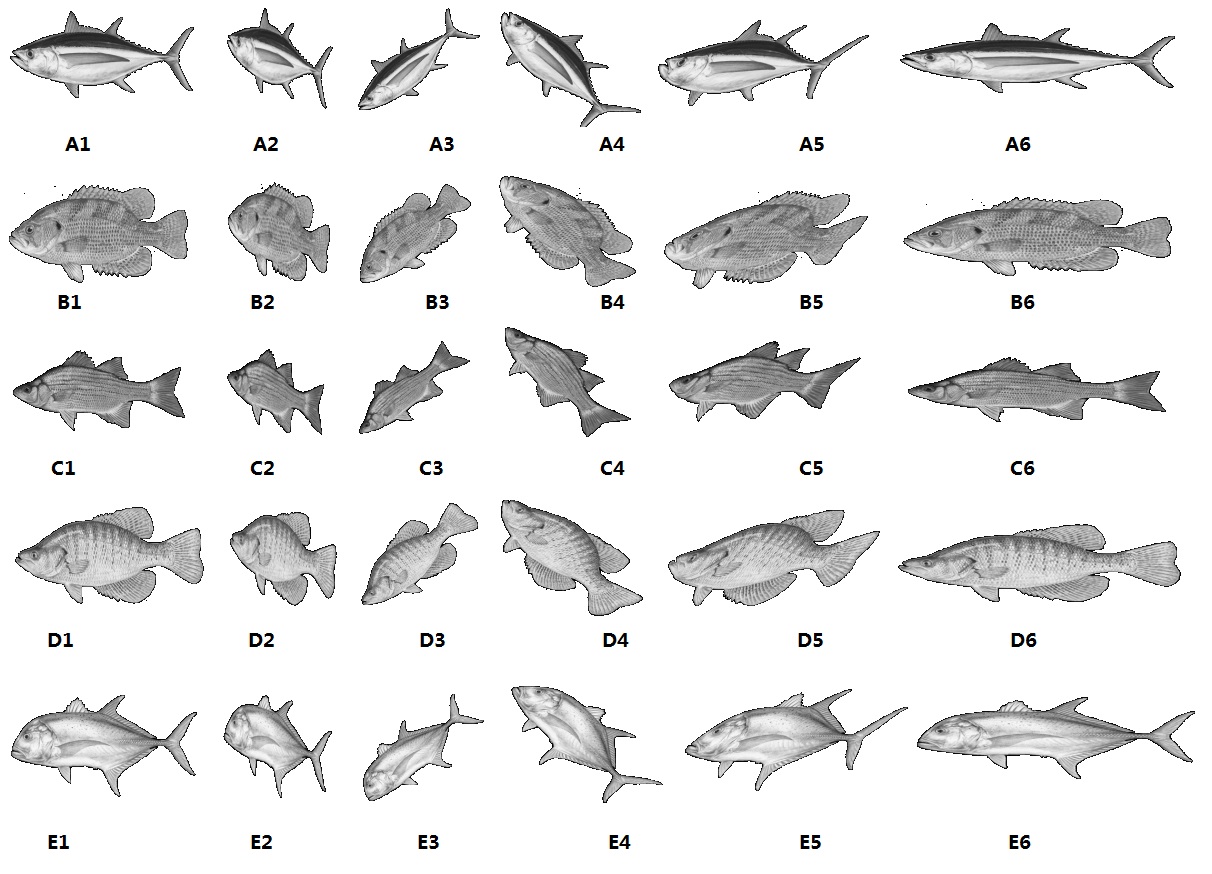}
  \caption{Test images, each line contains the original image and transformed versions.}\label{}
\end{figure}

AWMIs$\{AWMI^{1}_{1} \sim AWMI^{8}_{1}, AWMI_{2}\}$ are computed for each image in the Figure.1, and the results are presented in the Table 5. The error in the Table 5 is defined by
\begin{equation}
\hspace{-5mm}
error=\frac {Max(invariants)-Min(invariants)}{|Max(invariants)|+|Min(invariants)|}\times 100 \%
\end{equation}

According to the experimental results, we find that AWMIs have good stability and distinguishability. Therefore, it proved that the theoretical framework proposed in the previous sections is correct. Also, it's obviously that the error of $AWMI_{2}$ is greater those of others. This indicates that the error caused by the inaccuracy calculation of the high-order differentials will affect the performance of AWMIs.
\begin{table}[H]
\caption{The value of AWMIs}
\begin{tabular}{p{0.7cm}|p{1.3cm}|p{1.3cm}|p{1.3cm}|p{1.3cm}|p{1.3cm}|p{1.3cm}|p{1.3cm}|p{1.3cm}|p{1.3cm}c} 
\hline
& & & & & & & & &  \\
$No.$&$AWMI^{1}_{1}$&$AWMI^{2}_{1}$&$AWMI^{3}_{1}$&$AWMI^{4}_{1}$&$AWMI^{5}_{1}$&$AWMI^{6}_{1}$&$AWMI^{7}_{1}$&$AWMI^{8}_{1}$&$AWMI^{2}$  \\
& & & & & & & & &  \\
\hline
A1&$2.58e^{-5}$&$4.30e^{-9}$&0.0011&$1.70e^{-7}$&$-2.54e^{-6}$&$-5.43e^{-8}$&$1.10e^{-6}$&0.0138&-6.5376 \\
A2&$2.53e^{-5}$&$4.34e^{-9}$&0.0011&$1.74e^{-7}$&$-2.47e^{-6}$&$-5.24e^{-8}$&$1.10e^{-6}$&0.0138&-7.1118 \\
A3&$2.42e^{-5}$&$4.20e^{-9}$&0.0011&$1.72e^{-7}$&$-2.38e^{-6}$&$-4.85e^{-8}$&$1.08e^{-6}$&0.0135&-6.6618 \\
A4&$2.50e^{-5}$&$4.41e^{-9}$&0.0011&$1.79e^{-7}$&$-2.42e^{-6}$&$-5.00e^{-8}$&$1.05e^{-6}$&0.0137&-7.7582 \\
A5&$2.54e^{-5}$&$4.46e^{-9}$&0.0011&$1.79e^{-7}$&$-2.46e^{-6}$&$-5.15e^{-8}$&$1.06e^{-6}$&0.0138&-7.7967 \\
A6&$2.57e^{-5}$&$4.24e^{-9}$&0.0011&$1.67e^{-7}$&$-2.59e^{-6}$&$-5.43e^{-8}$&$1.10e^{-6}$&0.0137&-6.1510 \\
\hline
error &3.25\% &2.98\% &1.59\% &3.45\% &3.54\% &5.68\% &2.29\% &0.91\% &11.80\%  \\
\hline
B1&$2.54e^{-5}$&$7.45e^{-10}$&0.0010&$3.10e^{-8}$&$-2.60e^{-7}$&$-7.01e^{-9}$&$-5.25e^{-7}$&0.0122&0.3408 \\
B2&$2.52e^{-5}$&$7.38e^{-10}$&0.0010&$3.10e^{-8}$&$-2.54e^{-7}$&$-6.82e^{-9}$&$-5.18e^{-7}$&0.0121&0.3294 \\
B3&$2.44e^{-5}$&$7.10e^{-10}$&0.0010&$3.00e^{-8}$&$-2.44e^{-7}$&$-6.40e^{-9}$&$-5.17e^{-7}$&0.0120&0.3349 \\
B4&$2.50e^{-5}$&$7.66e^{-10}$&0.0010&$3.24e^{-8}$&$-2.54e^{-7}$&$-6.94e^{-9}$&$-5.21e^{-7}$&0.0121&0.3343 \\
B5&$2.52e^{-5}$&$7.91e^{-10}$&0.0010&$3.33e^{-8}$&$-2.63e^{-7}$&$-7.25e^{-9}$&$-5.24e^{-7}$&0.0121&0.3421 \\
B6&$2.53e^{-5}$&$7.32e^{-10}$&0.0010&$3.04e^{-8}$&$-2.60e^{-7}$&$-6.96e^{-9}$&$-5.25e^{-7}$&0.0122&0.3535 \\
\hline
error &2.04\% &5.45\% &1.04\% &5.11\% &3.86\% &6.23\% &0.79\% &0.68\% &3.52\%  \\
\hline
C1&$2.38e^{-5}$&$1.72e^{-9}$&0.0012&$8.62e^{-8}$&$-1.27e^{-6}$&$-2.62e^{-8}$&$-4.93e^{-7}$&0.0108&1.9451 \\
C2&$2.37e^{-5}$&$1.65e^{-9}$&0.0012&$8.29e^{-8}$&$-1.22e^{-6}$&$-2.51e^{-8}$&$-4.83e^{-7}$&0.0108&1.7535 \\
C3&$2.27e^{-5}$&$1.58e^{-9}$&0.0012&$8.05e^{-8}$&$-1.18e^{-6}$&$-2.34e^{-8}$&$-4.75e^{-7}$&0.0107&1.7962 \\
C4&$2.35e^{-5}$&$1.77e^{-9}$&0.0012&$8.93e^{-8}$&$-1.29e^{-6}$&$-2.66e^{-8}$&$-4.95e^{-7}$&0.0108&2.0037 \\
C5&$2.39e^{-5}$&$1.84e^{-9}$&0.0012&$9.24e^{-8}$&$-1.34e^{-6}$&$-2.80e^{-8}$&$-5.02e^{-7}$&0.0108&2.1434 \\
C6&$2.38e^{-5}$&$1.70e^{-9}$&0.0012&$8.54e^{-8}$&$-1.26e^{-6}$&$-2.57e^{-8}$&$-4.95e^{-7}$&0.0108&2.0168 \\
\hline
error &2.45\% &7.55\% &1.23\% &6.92\% &6.47\% &8.89\% &2.71\% &0.72\% &10.00\%  \\
\hline
D1&$2.29e^{-5}$&$5.30e^{-10}$&$8.54e^{-4}$&$1.92e^{-8}$&$-3.65e^{-7}$&$-9.92e^{-9}$&$-9.37e^{-8}$&0.0140&3.5345 \\
D2&$2.28e^{-5}$&$5.13e^{-10}$&$8.52e^{-4}$&$1.86e^{-8}$&$-3.59e^{-7}$&$-9.72e^{-9}$&$-9.26e^{-8}$&0.0140&3.4078 \\
D3&$2.21e^{-5}$&$4.91e^{-10}$&$8.39e^{-4}$&$1.81e^{-8}$&$-3.48e^{-7}$&$-9.28e^{-9}$&$-9.26e^{-8}$&0.0139&3.5029 \\
D4&$2.26e^{-5}$&$5.44e^{-10}$&$8.48e^{-4}$&$1.98e^{-8}$&$-3.62e^{-7}$&$-9.80e^{-9}$&$-9.28e^{-8}$&0.0140&3.3824 \\
D5&$2.28e^{-5}$&$5.71e^{-10}$&$8.53e^{-4}$&$2.07e^{-8}$&$-3.72e^{-7}$&$-1.01e^{-8}$&$-9.43e^{-8}$&0.0140&3.3863 \\
D6&$2.28e^{-5}$&$5.19e^{-10}$&$8.52e^{-4}$&$1.87e^{-8}$&$-3.59e^{-7}$&$-9.74e^{-9}$&$-9.47e^{-8}$&0.0140&3.6232 \\
\hline
error &1.84\% &7.52\% &0.92\% &6.92\% &3.32\% &4.38\% &1.12\% &0.55\% &3.44\%  \\
\hline
E1&$2.41e^{-5}$&$1.16e^{-9}$&$8.73e^{-4}$&$4.15e^{-8}$&$-9.91e^{-7}$&$-2.76e^{-8}$&$3.13e^{-7}$&0.0149&-5.3282 \\
E2&$2.37e^{-5}$&$1.17e^{-9}$&$8.67e^{-4}$&$4.21e^{-8}$&$-9.49e^{-7}$&$-2.62e^{-8}$&$3.13e^{-7}$&0.0148&-6.3237 \\
E3&$2.30e^{-5}$&$1.16e^{-9}$&$8.52e^{-4}$&$4.27e^{-8}$&$-9.17e^{-7}$&$-2.47e^{-8}$&$3.05e^{-7}$&0.0147&-5.8943 \\
E4&$2.34e^{-5}$&$1.23e^{-9}$&$8.61e^{-4}$&$4.49e^{-8}$&$-1.00e^{-6}$&$-2.76e^{-8}$&$3.11e^{-7}$&0.0148&-6.7204 \\
E5&$2.37e^{-5}$&$1.23e^{-9}$&$8.66e^{-4}$&$4.47e^{-8}$&$-1.02e^{-6}$&$-2.83e^{-8}$&$3.09e^{-7}$&0.0148&-6.6155 \\
E6&$2.41e^{-5}$&$1.15e^{-9}$&$8.73e^{-4}$&$4.14e^{-8}$&$-1.01e^{-6}$&$-2.78e^{-8}$&$3.12e^{-7}$&0.0149&-4.8477 \\
\hline
error &2.50\% &3.23\% &1.24\% &4.08\% &5.41\% &6.73\% &1.17\% &0.60\% &16.19\%  \\
\hline
\end{tabular}
\end{table}
\subsection{Image retrieval on ALOI dataset}
In this section, the performance of AWMIs is tested on ALOI(Amsterdam Library of Object Images) \cite{20}, which contains images of 1000 objects taking from 72 different viewpoints. All images in this dataset have black background. We choose 1200 images (size of $192 \times 144$) of 100 objects. Each object contains 12 images, which are shown in the Figure.2.
\begin{figure}
  \centering
  \includegraphics[width=10cm]{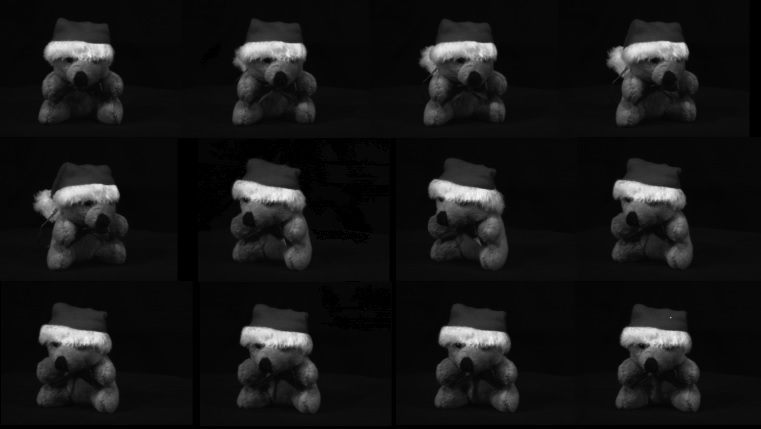}
  \caption{One object contains 12 images, which are taken from different viewpoints.}\label{}
\end{figure}
For comparison, we choose Hu-moments proposed in \cite{1}, AMIs proposed in \cite{6}, Zernike moments proposed in \cite{6} and Gaussian-Hermite moments \cite{15}.
For image retrieval, we adopt the modified $\chi^{2}$ distance to measure the similarity of two feature vectors \cite{21}, which is defined by

\begin{equation}
MD_{\chi^{2}}(V_{1},V_{2})=\frac {1}{n}\sum^{n}_{i=1}\frac {|v^{1}_{i}-v^{2}_{i}|}{|v^{1}_{i}|+|v^{2}_{i}|}
\end{equation}
where $V_{1}=(v^{1}_{1},v^{1}_{2},...,v^{1}_{n})$ and $V_{2}=(v^{2}_{1},v^{2}_{2},...,v^{2}_{n})$. We retrieval each image and draw 5 Precision-Recall curves of AWMIS, AMIs, HMs, Zernike moments and Gaussian-Hermite moments in Figure.3 to reflect their average levels of image retrieval.
\begin{figure}
  \centering
  \includegraphics[width=9cm]{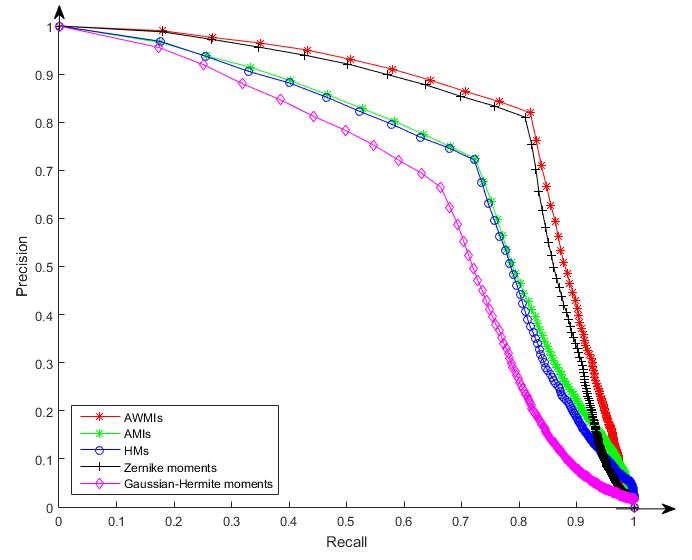}\\
  \caption{Precision-Recall curves of AWMIS, AMIs, HMs, Zernike moments and Gaussian-Hermite moments }\label{}
\end{figure}

Precision and Recall are defined by
\begin{equation}
Precision=\frac {| \left\{ relevant\ images \right\} \cap \left\{ retrieved\ images \right\} |}{|\left\{ retrieved\ images \right\}|}
\end{equation}
\begin{equation}
Recall=\frac {| \left\{ relevant\ images \right\} \cap \left\{ retrieved\ images \right\} |}{|\left\{ relevant\ images \right\}|}
\end{equation}

According Figure.3, it has been proved that, as a kind of global feature, AWMIs have better performance than other methods, such as AMIs. Because AWMIs combine local features with global features and contain more information of images. Meanwhile, orthogonal moments are invariant to the similarity transformation. So, when the affine transformation is applied to the image, retrieval performances of these moments will decrease. Finally, these new AWMIs greatly increased the number of low-order invariants. It should be noted that we can achieve better result of image retrieval by combining AWMIs with AMIs.

\section{Conclusion}
The contributions of this paper mainly include two aspects. Firstly, we extend the definition of moments,which is named as DMs. In theory, we can construct DMs containing arbitrary order partial derivatives. Secondly, by using local differential invariants and the structural framework of global integral invariants, we construct AWMIs. This approach greatly expands the number of low-order affine moments invariants. Meanwhile, it's important to note that there are many different ways to construct AWMIs. Thirdly, the final experimental results show that AWMIs have good stability and distinguishability. They also have better performance for image retrieval, too.

In the future, we will design more structural formulas to expand the number of AWMIs. At the same time, it's also important to explore the method of improving the accuracy of differential calculation, so that high-order differential moments can be used. Also, we want to combine the extraction method of AWMIs with deep-learning, so that the deep-learning network has invariance for some geometric transformations.

\end{document}